\documentclass{article}
\usepackage{spconf,amsmath,graphicx}
\usepackage{slashbox,pict2e}

\newcommand{\overbar}[1]{\mkern 1.5mu\overline{\mkern-1.5mu#1\mkern-1.5mu}\mkern 1.5mu}

\title{Experiments on Parallel Training of Deep Neural Network using Model Averaging}
\name{Hang Su$^{1,2}$, Haoyu Chen$^1$, Haihua Xu$^3$}
\address{$^1$ International Computer Science Institute, Berkeley, California, US \\
$^2$ Dept. of Electrical Engineering \& Computer Science, University of California, Berkeley, CA, USA \\
$^3$ Nanyang Technological University, Singapore \\
{\small \tt \{suhang3240@gmail.com\}}
}
\begin{document}
%\ninept
%
\maketitle
\begin{abstract}
In this work we apply model averaging to parallel training of deep neural network (DNN). 
Parallelization is done in a model averaging manner. Data is partitioned and distributed to different nodes for 
local model updates, and model averaging across nodes is done every few minibatches. 

We use multiple GPUs for data parallelization, and Message Passing Interface (MPI) for communication between nodes,
which allows us to perform model averaging frequently without losing much time on communication.
We investigate the effectiveness of Natrual Gradient Stochasitc Gradient Descent (NG-SGD) and Restricted Boltzmann 
Machine (RBM) pretraining for parallel training in model-averaging framework, and explore the best setups in 
term of different learning rate schedules, averaging frequencies and minibatch sizes. It is shown that NG-SGD 
and RBM pretraining benefits parameter-averaging based model training. On the 300h Swithboard dataset, a 9.3 times 
speedup is achieved using 16 GPUs and 17 times speedup using 32 GPUs with limited decoding accuracy loss. 
\footnote{This work is not submitted to 
peer-review conferences because the authors think it needs more investigation. The authors are in lack of 
resources to perform further exploration. However, we welcome any comments and suggestions.}
\end{abstract}
\section{Introduction}
\label{sec:intro}
Deep Neural Networks (DNN) has shown its effeciveness in several machine learning tasks, espencially in speech
recognition. The large model size and massive training examples make DNN a powerful model for classification. However,
these two factors also slow down the training procedure.

Parallelization of DNN training has been a popular topic since the revival of neural networks. Several different strategies
have been proposed to tackle this problem. Multiple thread CPU parallelization and single GPU implementation are compared
in \cite{scanzio2010parallel,vesely2010parallel}, and it is shown that single GPU could beat multi-threaded CPU implementation
by a factor of 2.

Optimality for parallelization of DNN training was analyzed in \cite{seide2014parallelizability}, and based on the analysis, 
a gradient quantization approach (1-bit SGD) was proposed to minimize communication cost \cite{seide20141}. It shows that 1 bit
quantization can effectively reduce data exchange in an MPI framework, and a 10 times speed-up is achieved using 40 GPUs.

DistBelief proposed in \cite{dean2012large} reports that 8 CPU machines train 2.2 times faster than a single GPU machine on a
moderately sized speech model. Asynchronous SGD using multiple GPUs achieved a 3.2x speed-up on 4 GPUs \cite{zhang2013asynchronous}.

A pipeline training approach was propoased in \cite{chen2012pipelined} and a 3.3x speedup was achieved using 4 GPUs, but this
method does not scale beyond number of layers in the neural network.

A speedup of 6x to 14x was achieved using 16 GPUs on training convolutional neural networks \cite{coates2013deep}. In this approach,
each GPU is responsible for a partition of the neural network. This approach is more useful for image classification where 
local structure of the neural network could be exploited. For a fully connected speech model, a model partition approach 
may not be able to contribute as much.

Distributed model averaging is used in \cite{zhang2014improving,miao2014distributed}, and a further improvement 
is done using NG-SGD \cite{povey2014parallel}. In this approach, separate models are trained on multiple nodes using 
different partitions of data, and model parameters are averaged after each epoch. It is shown that NG-SGD can effectively improve
convergence and ensure a better model trained using the model averaging framework.

Our approach is mainly based on the NG-SGD with model averaging. We utilize multiple GPUs in neural networks training via 
MPI, which allows us to perform model averaging more frequently and efficiently. Unlike the other approach \cite{seide20141},
we do not use a warm-up phase where only single thread is used for model update. (Admittedly, this might lead to further improvement). 
In this work, we conduct a lot of experiments and compare different setups in model averaging framework.

In Section 2, we introduce related works on NG-SGD. Section 3 describe the model averaging approach and some 
intuition on the analysis. Section 4 records experimental results on different setups and Section 5 concludes.

\section{Relationship to Prior Works}
To avoid confusion, we should mention that Kaldi\cite{kaldi11} contains two neural network recipes. The first implementation
\footnote{Location in code: src/\{nnet,nnetbin\}} is described in \cite{vesely2013sequence} which supports Restricted Boltzmann Machine 
pretraining \cite{hinton2006fast} and sequence-discriminative training \cite{povey2008boosted}. It uses single GPU 
for SGD training. The second implementation \footnote{Location in code: src/{nnet2,nnet2bin}}
\cite{zhang2014improving} was originally designed to support parallel training on multiple CPUs. Now it also supports 
multiple GPUs for training using model averaging. By default, it uses layer-wise discriminative pretraining.

Our work extends the first implementation so that it can utilize multiple GPUs using model averaging. We use MPI 
in implementation, so file I/O is avoided during model averaging. This allows us to perform model averaging much
more frequently. 

\section{Data parallelization and Model Averging}
SGD is a popular method for DNN training. Even though neural network training objectives are usually non-convex, 
mini-batch SGD has been shown to be effective for optimizing the objective\cite{seide2011conversational}. 
Roughly speaking, a bigger minibatch size gives a better estimate of the gradient, resulting in a better the converge rate. 
Thus, a straight forward idea for parallellization would be distributing the gradient computation to different computing
nodes. In each step, gradients of minibatches on different nodes are reduced to a single node, averaged and then used to 
update models in each node. This method, i.e. gradient averaging, can compute the gradient accurately, but it requires 
heavy communication between nodes.
Also, it is shown that increasing minibatch size does not always benefit model training\cite{seide2011conversational}, 
especially in early stage of model training.

On the other hand, if we choose to average the parameters rather than gradients, it is not necessary to exchange data that often. 
Currently, there is no straight forward theory that guarantees convergence, but we would like to explore a bit why this strategy 
should work, just as we observe in the experiments.

First, in the extreme case where model parameters are averaged after each weight update, model averaging is equivalent to 
gradient averaging. Furthermore, if model averaging is done every $n$ minibatch based weight update, model update formula could
be written as
\begin{equation}
\begin{split}
\theta_{t+n} &= \theta_{t} +\sum_{i=0}^{n-1} \alpha g_{t+i}\\
&= \theta_{t} +\sum_{i=0}^{n-1} \alpha \frac{\partial}{\partial\theta} F(x;\theta_{t+i})
\end{split}
\end{equation}
\begin{equation}
\overbar{\theta_{t+n}} = \overbar{\theta_{t}} +\sum_{i=0}^{n-1} \alpha \frac{\partial}{\partial\theta} \overbar{F}(x;\theta_{t+i})
\end{equation}
where $\theta$ is the model parameter and $\alpha$ is learning rate. If changes in model parameter $\theta$ is limited 
within $n$ updates, this approach could be seen as an approximation to gradient averaging.

Second, it is shown that model averging for convex models is guaranteed to converge \cite{mcdonald2010distributed,mcdonald2009efficient}.
It is suggested that unsupervised pretraining guides the learning towards basins of attraction of minima that support better generalization
from the training data set; \cite{erhan2010does}.

\begin{figure}[htb]
  \centering
  \includegraphics[width=0.5\textwidth]{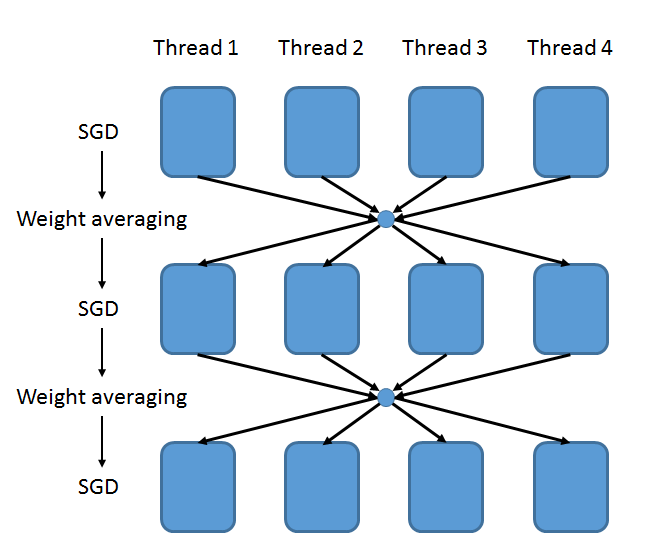}
  \caption{All-reduce network}
  \label{fig:allreduce}
\end{figure}

Fig~\ref{fig:allreduce} is an example of all-reduce with 4 nodes. This operation could be easily implemented by MPI\_Allreduce.

\section{Natural Gradient for Model Update}
This section introduces the idea proposed in \cite{povey2014parallel}.

In stochastic gradient descent (SGD), the learning rate is often assumed to be a scalar $\alpha_t$ that may change over time,
the update formula for model parameters $\theta_{t}$ is
\begin{equation}
\theta_{t+1} = \theta_{t} + \alpha_t g_t
\end{equation}
where $g_t$ is the gradient.

However, according to Natural Gradient idea \cite{murata1999statistical,roux2008topmoumoute}, it is possible to replace 
the scalar with a symmetric positive definite matrix $E_t$, which is the inverse of the Fisher information matrix.
\begin{equation}
\theta_{t+1} = \theta_{t} + \alpha_t E_t g_t
\end{equation}

Suppose $x$ is the variable we are modeling, and $f(x;\theta)$ is the probability or likelihood of $x$ given parameters $\theta$, then the
Fisher information matrix $I(\theta)$ is defined as
\begin{equation}
  E\bigg[\bigg(\frac{\partial}{\partial\theta}\log f(x;\theta)\bigg) \bigg(\frac{\partial}{\partial\theta}\log f(x;\theta)\bigg)^\top\bigg]
\end{equation}

For large scale speech recognition, it is impossible to estimate Fisher information matrix and perform inversion, 
so it is necessary to approximate the inverse Fisher information matrix directly. Details about the theory and 
implementation of NG-SGD could be found in \cite{povey2014parallel}.

\section{Experimental Results}
\subsection{Setup}
In this work, we report speech recognition results on the 300 hour Switchboard conversational telephone speech task 
(Switchboard-1 Release 2). We use MSU-ISIP release of the Switchboard segmentations and transcriptions (date 11/26/02),
together with the Mississippi State transcripts2 and the 30Kword lexicon released with those transcripts. 
The lexicon contains pronunciations for all words and word fragments in the training data. We use the Hub5 ’00 data for
evaluation. Specifically, we use the  the development set and Hub5 ’01 (LDC2002S13) data as a separate test set.

The Kaldi toolkit\cite{kaldi11} is used for speech recognition framework. Standard 13-dim PLP feature,
together with 3-dim Kaldi pitch feature, is extracted and used for maximum
likelihood GMM model training. Features are then transformed using LDA+MLLT before SAT training.
After GMM training is done, a tanh-neuron DNN-HMM hybrid system is trained using the the 40-dimension 
transformed fMLLR (also known as CMLLR \cite{gales1996generation}) feature as input and GMM-aligned senones 
as targets. fMLLR is estimated in an EM fashion for both training data and test data. A trigram language model (LM) is trained 
on 3M words of the training transcripts only.

Work in this paper is built on top of the Kaldi nnet1 setup and the NG-SGD method introduced in nnet2 setup. 
Details of DNN training follows Section 2.2 in \cite{vesely2013sequence}. In this work, we use 6 hidden layers, where each 
hidden layer has 2048 neurons with sigmoids. Input layer is 440 dimension (i.e. the context of 11 fMLLR frames), 
and output layer is 8806 dimension. Mini-batch SGD is used for backpropagation and the minibatch is set to 1024 for all
the experiments. By defult, DNNs are initialized with stacked restricted Boltzmann machines (RBMs) that are pretrained 
in a greedy layerwise fashion \cite{hinton2006fast}. Comparison between random initialization and RBM-initialization 
in model averaging framework is reported in Section~\ref{sec:init}.

The server hardware used in this work is Stampede (TACC) (URL: https://portal.xsede.org/tacc-stampede). It is a Dell Linux 
cluster provided as an Extreme Science Engineering Discovery Environment (XSEDE) digital service by the Texas Advanced 
Computing Center (TACC). Stampede is configured with 6,400 Dell DCS Zeus compute nodes, the majority of which are configured
with two 2.7 GHz E5-2680 Intel Xeon (Sandy Bridge) processors and one Intel Xeon Phi SE10P coprocessor. 128 of the nodes are 
augmented with an NVIDIA K20 GPU and 8 GB of GDDR5 memory each, which we use for neural network training in this work.
Stampede nodes run Linux 2.6.32 with batch services managed by the Simple Linux Utility for Resource Management (SLURM).

\subsection{Switchboard Results}
\begin{figure}[htb]
  \centering
  \includegraphics[width=0.5\textwidth]{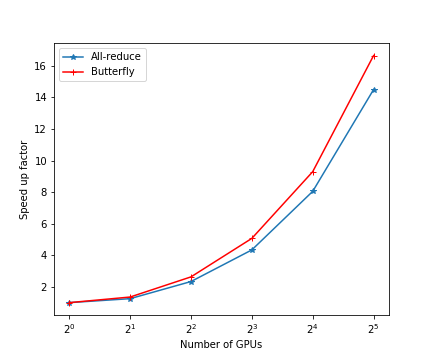}
  \caption{Speedup factor v.s. number of gpus}
  \label{fig:scaling}
\end{figure}

Fig.~\ref{fig:scaling} shows the speedup plot for model averaging experiments. As is shown in the graph,
a speedup of 17 could be achieved when 32 GPUs are used.
Table~\ref{tab:wer} shows the main decoding results for DNNs trained using different number of GPUs. In general, decoding
results of DNNs trained model averaging degrades 0.3\~0.4 WER, depending on the number of GPUs used.
\begin{table}
  \centering
  \begin{tabular}{|c|c|c|c|c|c|c|}
    \hline
    \backslashbox{Data}{Nodes}  & 1    & 2    & 4   & 8      & 16     & 32\\
    \hline
    SWB     & 14.7 & -- & --  & 15.1   & 15.1   & 15.2\\
    CallHM  & 26.8 & -- & --  & 27.4   & 27.0   & 27.1 \\
    \hline
    SWB     & 16.1  & -- & -- & 16.4  & 16.2  & 16.4 \\
    SWB2P3  & 21.0  & -- & -- & 21.8  & 21.7  & 21.7 \\
    SWB-Cell & 27.4 & -- & -- & 27.3  & 27.4  & 27.8 \\
    \hline
  \end{tabular}
  \caption{Comparison of WERs using different number of GPUs}
  \label{tab:wer}
\end{table}

\subsection{Initialization Matters}
\label{sec:init}
Table~\ref{tab:init} compares random initialization with Restricted Boltzmann Machine (RBM) based initialization.

\begin{table}
  \centering
  \begin{tabular}{c|c|c|c|c}
    \hline
    & \multicolumn{2}{c|}{SWB}  & \multicolumn{2}{c}{CallHome} \\
    \hline
    Nodes          &  1     & 32     &   1    & 32\\
    \hline
    random init     & 15.6  & 16.4   & 27.4   & 28.8 \\
    \hline
    RBM pretraining & 14.7  & 15.2   & 26.8   & 27.1 \\
    \hline
  \end{tabular}
  \caption{Comparing RBM pretraining with random initialization}
  \label{tab:init}
\end{table}
As we can see in the table, random initialization is worse than DNN with RBM pretraining by 0.9/0.6 in single GPU case.
While in model averaging setup, random initialization becomes even worse -- 0.3/0.9 point more degradation on WER.

\subsection{Averaging frequency}
Averaging frequency here is defined as the number of minibatch-SGD performed per model averaging.
Due to the limitation of computing resource, we only did preliminary experiments on this. Minibatch size of 1024 
is set as default, and we compare averaging frequency of 10 and 20. It is shown in Table~\ref{tab:freq} that an averaging 
frequency of 10 gives slight worse speedup but a better decoding WER. The tradeoff between lower averaging frequency 
(i.e. better speedup) and better training accuracy is within expectation in that frequent model averaging means 
steady gradient estimation.

\begin{table}
  \centering
  \begin{tabular}{c|c|c|c}
    \hline
    frequency   & Speedup   & SWB   & CallHome \\
    \hline
    baseline    &   --      & 14.7  & 26.8\\
    \hline
    10          & 9.32      & 15.1  & 27.0 \\
    \hline
    20          & 10.07     & 15.8  & 28.0 \\
    \hline
  \end{tabular}
  \caption{Comparing different averaging frequencies}
  \label{tab:freq}
\end{table}

\subsection{Minibatch Size}
Table~\ref{tab:mbsize} compares two different minibatch size in model averaging setup.
\begin{table}
  \centering
  \begin{tabular}{c|c|c|c|c|c}
    \hline
    & \multicolumn{2}{c|}{SWB}    & \multicolumn{2}{c|}{CallHome} & Speedup\\
    \hline
    nodes       &  1    &   16    &    1     &   16   &    \\
    \hline
    256         & 15.3     & 15.6 & 26.8     &  27.3   & -- \\
    \hline
    1024        & 14.7     & 15.1 & 26.8     &  27.0   & 9.32\\
    \hline
  \end{tabular}
  \caption{Comparing different minibatch size}
  \label{tab:mbsize}
\end{table}

\subsection{Learing Rate Schedule}
Initial learning rate is increased in porportion to number of threads in model averaging setup. The reason for this is straight 
forward:
Assume we have $n$ minibatches of data for model training. When the model is trained using single thread, it gets updated $n$
times. When data is distributed to $m$ machines, then each model gets updated $n/m$ times. Since the effect of model averaging 
is mostly aggregating knowledge learnt from different data partition, the absolute change of model shall be compensated by 
$m$ times.

We compare two learning rate schedules in this section. The first one is the default setup used in Kaldi nnet1 (Newbob). 
It starts with a initial learning rate of 0.32 and halves the rate when the improvement in frame accuracy on a cross-validation 
set between two successive epochs falls below 0.5\%. The optimization terminates when the frame accuracy increases by less 
than 0.1\%. Cross-validation is done on 10\% of the utterances that are held out from the training data.

The second learing rate schedule is exponentially decaying. This method is used in \cite{senior2013empirical,povey2014parallel} 
and is shown to be superior to performance scheduling and power scheduling. In this work, it starts with the same initial learning
rate as the first method (Newbob), and decrease to the final learning rate (which is set to be 0.01 * initial learning rate). The
number of epochs is set to $15$ in this task, which is set to be the same as Newbob scheduling.

As is shown in Table~\ref{tab:lr}, these two learning rate scheduling methods give similar decoding results. However, 
exponential learning rate might need more tuning since it requires a initial learning rate, a final learning rate and
predefined number of epochs to train.
\begin{table}
  \centering
  \begin{tabular}{c|c|c|c|c}
    \hline
    & \multicolumn{2}{c|}{SWB}  & \multicolumn{2}{c}{CallHome} \\
    \hline
    Nodes       &  1      & 16    &  1    & 16 \\
    \hline
    Newbob      & 14.9    & 15.4  & 26.6  & 27.2 \\
    \hline
    exponential & 14.7    & 15.1  & 26.8  & 27.0 \\
    \hline
  \end{tabular}
  \caption{Comparing learning rate schedule}
  \label{tab:lr}
\end{table}

\subsection{Online NG-SGD Matters}
Table~\ref{tab:ngsgd} compares plain SGD with NG-SGD in model averaging mode, and it shows NG-SGD is crucial to 
model training with parameter-averaging.
\begin{table}
  \centering
  \begin{tabular}{c|c|c|c|c}
    \hline
    & \multicolumn{2}{c|}{SWB}  & \multicolumn{2}{c}{CallHome} \\
    \hline
     Nodes  &  1    & 16        &   1   &  16 \\
    \hline
    SGD     & 14.9    & 16.3    &  26.9 & 28.3 \\
    \hline
    NG-SGD  & 14.7    & 15.1    &  26.8 & 27.0  \\
    \hline
  \end{tabular}
  \caption{Comparing NG-SGD and naive SGD}
  \label{tab:ngsgd}
\end{table}

\section{Conclusion and Future Work}
In this work, we show that neural network training can be efficiently speeded up using model averaging. 
on a 300h Switchboard dataset, a 9.3x / 17x speedup could be achieved using 16 / 32 GPUs respectively, 
with limited decoding accuracy loss. We also show that model averaging benefits a lot from NG-SGD and RBM based pretraining.
Preliminary experiments on minibatch size, averaging frequency and learning rate schedules are also presented.

Further accuracy improvement might be achieved if parallel training runs on top of serial training initialization.
It would be interesting to see if sequence-discriminative training combines well with model averaging.
Speedup factor could be further improved if CUDA aware MPI is used. Theory on convergence using model averaging is 
to be explored, which might be useful for guiding future development.

\section{Acknowledgements}
We would like to thank Karel Vesely and Daniel Povey who wrote the original "nnet1" neural network training code
and natural gradient stochastic gradient descent upon which the work here is based. We would also like to thank Nelson 
Morgan, Forrest Iandola and Yuansi Chen for their helpful suggestions. 

We acknowledge the Texas Advanced Computing Center (TACC) at The University of Texas at Austin for providing 
HPC resources that have contributed to the research results reported within this work (URL: http://www.tacc.utexas.edu).

% References should be produced using the bibtex program from suitable
% BiBTeX files (here: strings, refs, manuals). The IEEEbib.bst bibliography
% style file from IEEE produces unsorted bibliography list.

% -------------------------------------------------------------------------
\bibliographystyle{IEEEbib}
\bibliography{paper}

\end{document}